\documentclass[10pt, a4paper]{article}

\usepackage[final]{lrec2026}
\usepackage{xcolor}
\usepackage{booktabs}
\usepackage{enumitem}
\usepackage[table]{xcolor}
\usepackage{graphicx}
\usepackage{multirow}

\title{LLM-Based Data Generation and Clinical Skills Evaluation for Low-Resource French OSCEs}

\name{Tian Huang, Tom Bourgeade, Irina Illina} 

\address{Universit\'e de Lorraine, CNRS, Inria, LORIA, \\
         F-54000 Nancy, France \\
         tian.huang@loria.fr\textsuperscript{$\dagger$}, tom.bourgeade@loria.fr, irina.illina@loria.fr\\}

\abstract{
Objective Structured Clinical Examinations (OSCEs) are the standard method for assessing medical students’ clinical and communication skills through structured patient interviews. In France, however, the organization of training sessions is limited by human and logistical constraints, restricting students’ access to repeated practice and structured feedback.
Recent advances in Natural Language Processing (NLP) and Large Language Models (LLMs) now offer the opportunity to automatically evaluate such medical interviews, thereby alleviating the need for human examiners during training. Yet, real French OSCE annotated transcripts remain extremely scarce, limiting reproducible research and reliable benchmarking.
To address these challenges, we investigate the use of LLMs for both generating and evaluating French OSCE dialogues in a low-resource context. We introduce a controlled pipeline that produces synthetic doctor–patient interview transcripts guided by scenario-specific evaluation criteria, combining ideal and perturbed performances to simulate varying student skill levels. The resulting dialogues are automatically silver-labeled through an LLM-assisted framework supporting adjustable evaluation strictness.
Benchmarking multiple open-source and proprietary LLMs shows that mid-size models ($\le$32B parameters) achieve accuracies comparable to GPT-4o ($\sim$90\%) on synthetic data, highlighting the feasibility of locally deployable, privacy-preserving evaluation systems for medical education.
 \\ \newline }

\begin{document}
\maketitleabstract
\begingroup
\renewcommand{\thefootnote}{$\dagger$}
\footnotetext{Work conducted while the author was an intern at LORIA. The institutional email is no longer active; please use thuang44@gmail.com for correspondence.}
\endgroup

\section{Introduction}

Effective clinical and communication skills are essential in healthcare practice, where doctor–patient interviews form the foundation of diagnosis, treatment, and patient trust. However, training opportunities for these skills remain limited, primarily due to the dependence on human participants, which increases costs and reduces accessibility.
Recent advances in Natural Language Processing now make it feasible to automate key aspects of this training process, including the generation of realistic medical dialogues and the automated evaluation of student performance.

Objective Structured Clinical Examinations (OSCEs) have become the de facto international standard for assessing medical students through simulated encounters. In the French implementation (\textit{Examens Cliniques Objectifs Structurés, ECOS}), a student plays the role of a doctor interacting with a trained actor serving as a standardized patient (SP), under the observation of an examiner who assesses the student’s clinical and communication skills, during 7-10 min pre-defined scenarios, referred to as ``\textbf{stations}''. Only a few French medical schools can organize up to two weekly OSCE training sessions for final-year students preparing for the national exams. However, not all students can attend every session, as limited logistical and human resources restrict the number of available slots.

The ability to perform repeated training with structured feedback would thus significantly benefit students’ preparation, and as such, automated evaluation systems leveraging advances in Natural Language Processing and Large Language Models (LLMs) may offer a potential solution. In this work, we focus in particular on the \textbf{automated evaluation of clinical skills in doctor-patient dialogue transcripts} (as could be obtained from a speech-to-text model, though we focus on text only here), which could alleviate the workload of, or even completely replace, human examiners in training sessions.

Recent studies have demonstrated the feasibility of LLM-based evaluation of English OSCE transcripts, showing high levels of agreement with human examiners across specific criteria \citep{shakur2024largelanguagemodelsmedical,geathers2025benchmarkinggenerativeaiscoring}. However, these works are grounded in the English OSCE tradition, where evaluation relies on standardized, station-agnostic checklists. In contrast, French OSCEs employ highly station-specific and heterogeneous criteria, making direct adaptation of such methods more challenging. Additionally, data for OSCEs, annotated or otherwise, remains extremely scarce, further limiting research and reproducibility in this area. In particular, no public French corpus of OSCE text transcripts has been made available to date.
Other research has explored synthetic corpora of medical dialogues as a way to overcome data scarcity  \citep{Wang_2024,das_SyntheticPatientPhysicianDialogue_2024}. While these efforts have yielded large-scale resources, they primarily cover English and Chinese, or rely on standardized country-specific structures for clinical interview notes. In contrast, the French OSCE context remains largely low-resource, with scarce annotated data, and no country-wide standardized structure for clinical interviews and notes.

To address these challenges of data scarcity and high variety in evaluation criteria, we propose leveraging LLMs to generate synthetic French OSCE doctor–patient transcripts from existing training scenarios and to automatically evaluate them against their associated clinical‑skills checklists. Specifically, our main contributions are as follows: (1) A controlled pipeline is presented for generating French OSCE doctor-patient dialogues, capturing both ideal and suboptimal student performances. We also propose an LLM-assisted labeling framework offering adjustable levels of strictness; (2) several prompting strategies are examined, along with two auxiliary tools designed to support smaller, locally hostable LLMs ($\leq$ 32B parameters) in evaluating transcripts against binary criteria checklists; (3) a benchmarking study across a range of open‑source and proprietary LLMs provides an initial assessment of the feasibility of this task in this low‑resource educational context.

\section{Related Work}

\textbf{AI for Pedagogical Assessment}
has been increasingly applied in education, with demonstrated benefits for both learning outcomes and scalability. Studies have reported measurable improvements in student outcomes and efficiency, including reductions in examiner workload \citep{Alizadeh2025}. Simulation-based education has also seen the deployment of AI-driven scoring systems that provide large-scale feedback and better return on investment compared to traditional human-based evaluations \citep{Campbell2025}. Automated short-answer grading tools have shown strong correlations with human examiners \citep{Seneviratne2025}. These advances establish the feasibility of AI for non-interactive, text-based evaluation tasks.

\textbf{LLMs for OSCE Evaluation}
have started to be explored as a means of automatically assessing medical students clinical and communication skills. \citet{Jamieson2024} reframed OSCEs criteria as prompts for evaluating post-dialogue clinical notes, while \citet{shakur2024largelanguagemodelsmedical} demonstrated high agreement ($\kappa = 0.88$) between GPT-4 and human examiners on single criterion, in recorded video transcripts. \citet{geathers2025benchmarkinggenerativeaiscoring} extended this line of work to multi-criteria scoring across 28 generic items(\href{https://health.uconn.edu/principles-clinical-medicine-clinical-skills-assessment/master-interview-rating-scale-mirs/}{MIRS evaluation rubrics}), testing several prompting strategies. Collectively, these studies confirm the feasibility of transcript-based evaluation, but they remain limited to English datasets and standardized criteria. The adaptation to French OSCEs is particularly challenging due to heterogeneous, station-specific criteria and the lack of annotated corpora.

\textbf{Calibration and Bias in LLM-based Assessment}
remain key challenges: while LLMs offer scalability and reduced examiner fatigue \citep{Haider_Badshah_Khan_Ullah_2018}, they also introduce variability, with studies documenting systematic differences in grading tendencies \citep{wei2025systematicevaluationllmasajudgellm} and biases inherited from training corpora \citep{santurkar2023opinionslanguagemodelsreflect,gallegos2024biasfairnesslargelanguage}. Surveys of ``LLM-as-a-judge'' approaches emphasize both their promise for large-scale evaluation and their sensitivity to prompt design and contextual shifts \citep{gu2025surveyllmasajudge}. These findings motivate the need for explicit calibration mechanisms, and robust validation before deployment in high-stakes educational settings.

\textbf{Synthetic Medical Dialogue Corpora}
 have been investigated as a means to overcome data scarcity. \citet{Wang_2024} introduced NoteChat, a multi-agent framework generating physician–patient interactions conditioned on clinical notes. SynDial \citep{das_SyntheticPatientPhysicianDialogue_2024} proposed iterative feedback loops to align generated dialogues with case-specific constraints. Large-scale corpora such as \textit{MTS-Dialog} \citep{ben-abacha-etal-2023-empirical} and \textit{MedDialog} \citep{zeng-etal-2020-meddialog} provide valuable resources, mainly in English and Chinese, though a French-translated variant, \textit{MedDialog-FR}, has been made available by \citet{liu_MedDialogFRFrenchVersion_2024}. \citet{nun_SIMSAMUFrenchMedical_2025} propose a large-scale dataset of transcriptions of simulated real-life medical dispatch calls by French junior emergency responders, alongside an expert-validated emergency dispatch dialog scheme. While these initiatives demonstrate feasibility, they do not capture the specificity of French OSCEs: heterogeneous, station-dependent evaluation criteria, which can sometimes be compositional, consisting of multiple sub-criteria linked by logical operators. This gap underlines the need for controlled synthetic corpora tailored to Francophone OSCE scenarios.

\section{Data Generation}
\label{sec:generation}
\subsection{LLM-Based Dialogue Generation}

\begin{figure*}[h]
\centering
\includegraphics[width=0.90\textwidth]{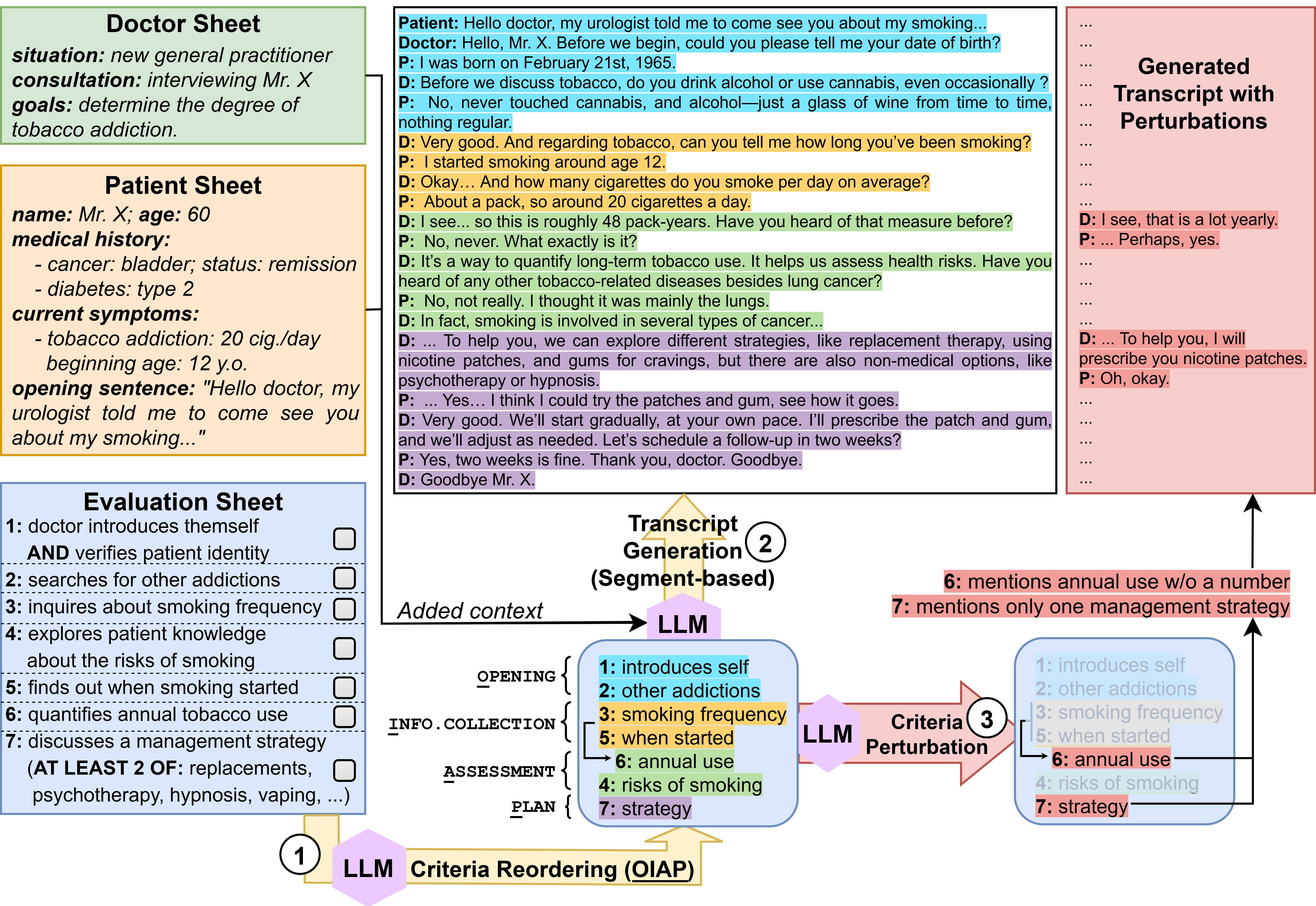}
\caption{Dialogue Generation pipeline. The Doctor/Patient/Evaluation Sheets are structured documents (JSON). \textbf{(1)} The list of binary clinical skills criteria is reordered via LLM, respecting the OIAP structure and presence of dependencies (\#6 here depends on \#3 and \#5). \textbf{(2)} A synthetic ideal-performance transcript is generated via LLM in segments using $N$ criteria slices ($N=2$ in this illustration). \textbf{(3)} To generate a less ideal transcript for data variety, some criteria are perturbed, leading to their failure in the generation.}
\label{fig:generation_pipeline}
\end{figure*}

To assess the feasibility of automating OSCE evaluation with LLMs, transcribed doctor–patient interactions are required, as they provide the textual input from which the model can infer, for each evaluation criterion, a binary outcome indicating whether it was met or not (see Fig.~\ref{fig:eval_pipeline}).
However, no public corpus of French OSCE transcripts exists, as recordings from both training and examination sessions are confidential and protected by data privacy regulations.

A robust dataset for this task should include examples covering a variety of medical student performance, from ideal to less well executed, matching the range of grades and comments that can be attributed by examiners. Yet obtaining examples of ideal performance is challenging: students seldom reach expert-level execution during training, and transcriptions of tutor- or faculty-led demonstrations, which could serve as references, are comparatively rare. LLM-based dialogue transcript generation thus offers a controlled alternative, allowing experimentation and reproducibility while avoiding the limitations of scarce or sensitive real-world data.

\textbf{Guiding Criteria:}
To make such generation feasible and coherent, we rely on the structured materials provided with each OSCE station, which describe the scenario from the perspectives of the doctor, patient, and examiner (see \autoref{fig:generation_pipeline}). Each French OSCE station comes with three documents that specify its scenario: a \textbf{doctor sheet}, which summarizes the situation, patient identity, and goals of the student during the interview; a \textbf{patient sheet}, which provides the background information, symptoms, and intended behavior of the patient; and an \textbf{evaluation sheet}, which contains a list of observable criteria (e.g., \textit{Inquires about the start of the symptoms}) that the doctor can pass or fail, and which are thus graded in a binary (\textit{done}/\textit{not done}) manner. These criteria are initially presented in an arbitrary order.

The corpus generation process used in this work follows the multi-step pipeline detailed in \autoref{fig:generation_pipeline}. Inspired by \citet{Wang_2024}, we propose using evaluation sheets, which typically contain around $\sim 20$ binary criteria, as a sort of ``script'' to guide the generation of the doctor's dialogue turns during the simulated interview. First, the list of binary criteria is reordered by an LLM to improve dialogue coherence and naturalness, preserving potential dependencies between criteria. Following, slices of this list are successively fed to a dialogue generation LLM, together with the doctor and patient sheets, to generate segments of the dialogue, which are finally concatenated to obtain a generated transcript of an OSCE station. Because data generation is a one-time process that directly influences corpus quality and entails no data privacy issues, we used an industry-leading large language model (GPT-4o) to ensure the overall quality and realism of the generated dialogues.

\textbf{Criteria Reordering:}
The order of these guiding criteria is critical, as they directly impact the order of generated speech turns and thus the naturalness and coherence of the synthetic dialogues.
We thus manually picked one of two different ordering strategies, based on the specific context of each OSCE station:
\begin{enumerate}[label=(\arabic*),itemsep=0pt,parsep=1pt,leftmargin=0pt,itemindent=*,topsep=0pt]
    \item \label{item:OIAP} the \textbf{OIAP structure} (\textit{Opening and Preparation, Information Collection, Assessment, Plan}), adapted from the SOAP framework \citep{podder_SOAPNotes_2023}, as operationalized by \citet{Wang_2024}, with modifications partly inspired by the Calgary-Cambridge model \citep{kurtz_CalgaryCambridgeReferencedObservation_1996} of medical interview structuring. In \citet{Wang_2024}, dialogues were generated in the SOAP order—\textit{Subjective} (patient-reported information), \textit{Objective} (clinically observable data), \textit{Assessment}, and \textit{Plan}—with each section produced separately and then concatenated. To better align with French OSCE training practices, we added an \textit{Opening and Preparation} phase covering mandatory preliminary actions such as verifying the patient’s identity, as mandated by the 2021 French regulation on patient identity vigilance \citep{MinSante_identitovigilance}. We also merged the \textit{Subjective} and \textit{Objective} parts into a single \textit{Information Collection} block to avoid unnatural dialogue breaks. For instance, when one evaluation criterion requires the student to inquire about smoking frequency (\textit{Subjective}) and another to calculate the pack-year value (\textit{Objective}), separating them would cause unnatural distance during generation. The OIAP definition is included in the prompt to guide the LLM in ordering the station’s evaluation criteria.
    \item a \textbf{context-driven ordering}, used for atypical cases where the OIAP sequence would be unnatural, such as when a patient insists from the beginning on being referred for bariatric surgery. In such situations, the doctor may need to explain the preconditions for surgery (which would normally fall under the \textit{Assessment} phase) rather than starting with the usual \textit{Opening} or \textit{Information Collection} phases. For these stations, the LLM is instructed through a different prompt to order the evaluation criteria directly based on the given doctor–patient context.
\end{enumerate}

\textbf{Criteria Perturbation:}
For our objective of assessing whether LLMs can reliably evaluate medical student performance in OSCEs, the generated corpus needs to include simulated doctor–patient interactions that reflect different levels of student performance. To obtain such diversity while maintaining dialogue coherence, we introduce a \textbf{perturbation} step in which a subset of the criteria guiding dialogue generation are replaced with \textbf{perturbed variants}, which deliberately distort their original intent, prompting the virtual doctor to perform incorrect or incomplete actions. The generated dialogue thus tends to fail the original unperturbed criteria, producing a suboptimal yet coherent doctor performance.
However, some criteria can have dependencies: for example, a doctor cannot quantify the annual consumption of tobacco without asking the patient their smoking frequency (see \autoref{fig:generation_pipeline}). We thus task \texttt{GPT-4o} with identifying what we refer to as \textit{leaf criteria}, that is, those whose correct execution does not depend on other criteria. 
Only these leaf criteria are then considered for perturbation. Perturbations enable the generation of more diverse data while preserving overall coherence in generated dialogue transcripts. Examples are illustrated in \autoref{fig:generation_pipeline}, e.g., an original criterion requiring the doctor to quantify the patient’s annual use of tobacco is perturbed such that providing a number is no longer required, resulting in failing the original criterion.

Once reordered, dialogue generation proceeds in sequential slices of $N$ criteria (we found $N=4$ provides good quality generated transcripts after preliminary experiments): for each slice, prompts combine station context (doctor and patient sheets), previous segments of the generated transcript, and the target criteria. The LLM then generates a new dialogue segment, which is concatenated with the previous ones. The criteria within a segment may not follow the slice's order, allowing for variations that enhance variety without affecting coherence (e.g., criteria \#3 and \#5 are swapped in \autoref{fig:generation_pipeline}).
Minor post-processing prompting is performed, namely on the first and final segments, to ensure dialogue openings and closings are present, improving naturalness.
From available training scenario material, we selected \textbf{10 representative OSCE stations} focused exclusively on doctor–patient dialogue, excluding those requiring external artifacts (use of medical devices, diagnostic maneuvers, or documents) or atypical criteria, resulting in a total of \textbf{179 binary criteria}. For each of these 10 stations, two synthetic dialogues were generated and grouped into: a \textbf{unperturbed} (optimal execution) corpus and a \textbf{perturbed} (with a chosen 50\% perturbation rate, determined through preliminary experiments) corpus.

Since the LLM treats the evaluation criteria as guidance for dialogue generation, the passing or failing of any specific criterion cannot be guaranteed. To confirm the influence of this guidance, after automatic silver-labeling (see \autoref{sec:labeling} below), we measured the proportion of failed criteria in both the \textbf{unperturbed} and \textbf{perturbed} corpora (\autoref{tab:false_ratio_comparison}). As expected, the \textbf{perturbed} dialogues showed a markedly higher proportion ($\sim40\%$) compared to \textbf{unperturbed} ($\sim10\%$).

\begin{table}[htbp]
\centering
\scriptsize
\begin{tabular}{lrrr}
\toprule
\textbf{Station ID} & \textbf{Criteria} & \textbf{Failed} \textbf{\textit{unperturbed}} & \textbf{Failed} \textbf{\textit{perturbed}} \\
\midrule
113 & 17 & 0.0\%  & 29.4\% \\
128 & 18 & 5.6\%  & 61.1\% \\
165 & 19 & 15.8\% & 52.6\% \\
\multicolumn{4}{c}{\ldots} \\
\midrule
\textbf{Total} & \textbf{179} & \textbf{10.6\%} & \textbf{39.7\%} \\
\bottomrule
\end{tabular}
\caption{Proportion of criteria annotated as \textit{failed} (not done) in the unperturbed and perturbed corpora}
\label{tab:false_ratio_comparison}
\end{table}

As the dialogue generation is constrained by the predefined list of evaluation criteria, the resulting synthetic doctor-patient interactions remain closely tied to those items and do not reflect the broader range of behaviors that real students might exhibit during OSCE encounters. This makes them more idealized than real OSCE interactions, where students sometimes deviate from the expected structure and display spontaneous or unexpected behaviors. Even with perturbed variants, the dialogues rarely deviate beyond the specified criteria. This tendency is further amplified by the inherent bias of large language models toward cooperative or sycophantic responses \cite{sharma_UnderstandingSycophancyLanguage_2023}, which often produce well-structured and overly agreeable dialogues. While this affects the realism of the generated data, it also offers advantages: using explicit checklists as generation guides creates a controlled and reproducible environment for our objective of determining the viability of LLM-as-evaluators in OSCEs. Furthermore, we hypothesize that if the evaluation pipeline remains consistently strict under such idealized conditions, it would demonstrate the robustness needed to provide fair and useful feedback to students, making it a valuable tool for helping them articulate their reasoning and communication more clearly during OSCE training.

\subsection{LLM-Assisted Silver-Labeling}\label{sec:labeling}

In addition to the scarcity of French OSCE transcripts, expert-completed evaluation sheets are also difficult to obtain. In practice, during training sessions, the examiner role is often played by other medical students, resulting in indicative rather than fully reliable assessments. To obtain reference evaluation labels for the generated dialogues, we therefore adopted an \textbf{LLM-assisted silver-labeling} approach, in which \texttt{GPT-4o} was tasked with producing preliminary binary labels for each evaluation criterion, along with a justification text and evidence extracted from the transcript to support subsequent manual verification. The resulting silver-labels were then reviewed manually and, when necessary, adjusted to ensure consistency with the intended evaluation standards, which are further detailed below.

During preliminary experiments, we observed that the strictness of the labeling prompt had a strong influence on the model’s decisions. When instructed to “strictly evaluate” whether a student explicitly performed an action, the model tended to assign fewer positive (corresponding to a student passing a criterion) silver labels than when using a more relaxed formulation such as “evaluate whether this criterion is met.” When the same data, model, and prompt were kept fixed, repeated runs with random seeds produced highly consistent results (Cohen’s $\kappa \approx 0.9$), indicating strong internal reliability. However, when only the prompt phrasing was changed from stricter to more relaxed, the agreement dropped substantially ($\kappa = 0.66$), showing that the model’s judgments were very sensitive to the level of evaluation rigor. Given these observations and the absence of expert-provided labels, we experimented with two explicit evaluation standards, referred to as \textbf{soft mode} and \textbf{strict mode}, to structure subsequent silver-labeling experiments.
In the \textbf{soft mode}, a criterion is considered passed (doned) once the target information appears in the dialogue, even if introduced by the patient rather than explicitly elicited or confirmed by the student. For example, the criterion “Inquires about smoking frequency” is considered passed if the patient spontaneously says, “I smoke ten cigarettes a day”, after a general question like, “Do you smoke?”. In the \textbf{strict mode}, the student must explicitly elicit or acknowledge the information, for example here, by acknowledging the notion of frequency explicitly with “That seems quite frequent.”

Though this labeling step is performed without direct expert involvement, the silver-labeling standards and modes of strictness used here were inspired by French OSCE redaction guidelines provided to us by OSCE organizers, and validated by real training sessions we witnessed. They provide a transparent and reproducible basis for assessing how reliably an LLM can identify and judge behaviors according to predefined criteria in doctor-patient interactions.
\section{Experimental Setup}
\vspace{-1.2em}
\begin{figure}[h]
\centering
\includegraphics[width=0.80\columnwidth]{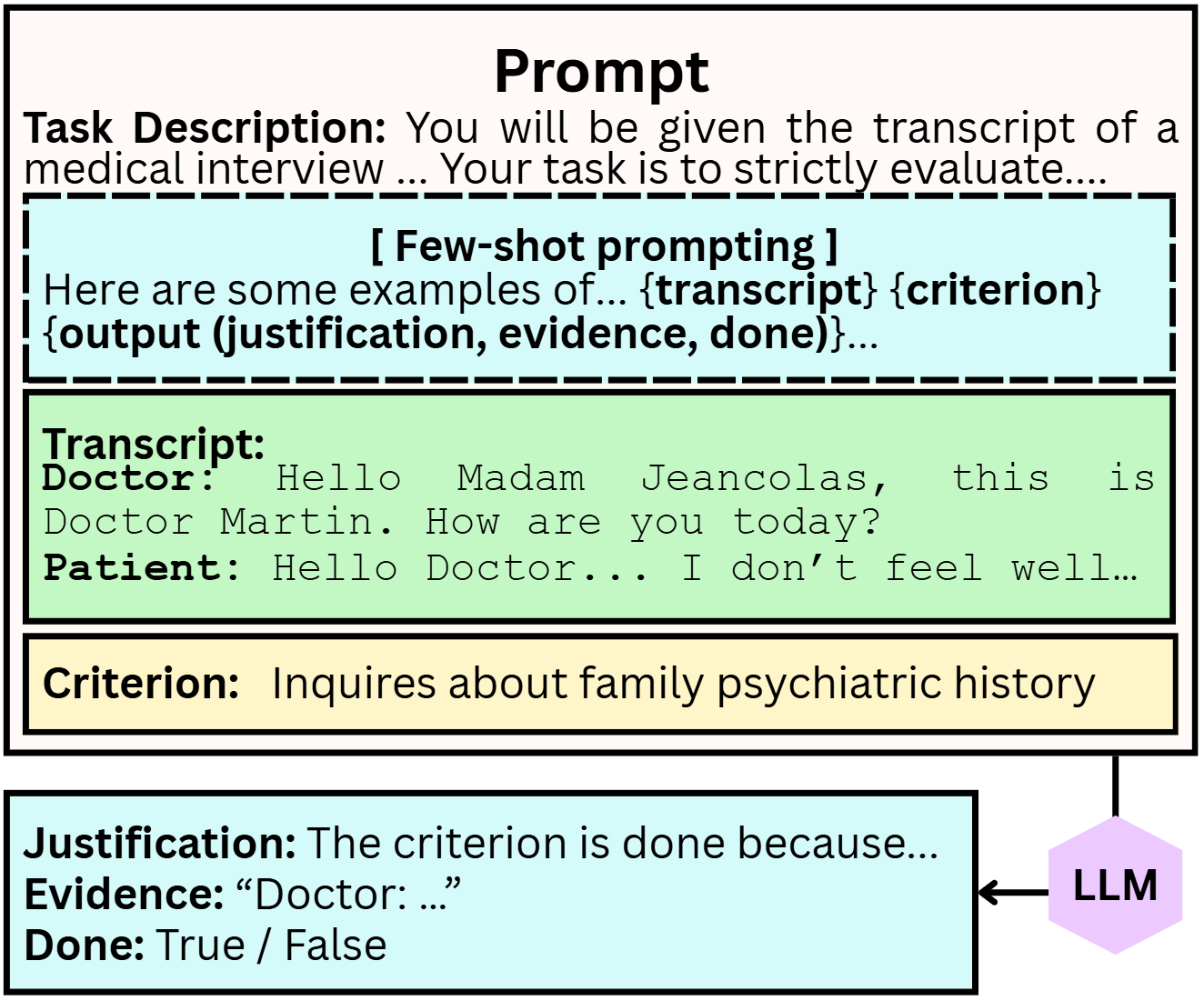}
\caption{Overview of the evaluation pipeline: an LLM is tasked with judging whether a specific criterion is passed/failed (\textit{Done}), with a \textit{justification} and supporting \textit{evidence}, given a transcript, and optionally some \textit{few-shot} examples.}
\label{fig:eval_pipeline}
\end{figure}

The overall evaluation workflow is illustrated in \autoref{fig:eval_pipeline}.
For each clinical case, the input to the evaluation system consists of the full \textbf{transcript}, a single \textbf{criterion}, and a default \textbf{task description} serving as the evaluation prompt. The task description instructs the LLM to strictly evaluate (similar to \textbf{strict mode}, see \autoref{sec:labeling}) the transcript against the given criterion and output the result in JSON format. Given these inputs, the LLM produces three outputs, in the following order: (i) a short \textbf{justification} paragraph; (ii) supporting \textbf{evidence}, consisting of the transcript segments most relevant to that decision; and (iii) a binary \textbf{true/false decision} indicating whether the criterion is satisfied. In addition to acting as explanations for an output, (i) and (ii) are intended to serve as a task-specific form of \textit{chain-of-thought} reasoning \citep{wei2022chain}, hence why they precede the binary label decision.

To assess the feasibility of LLM-based automated OSCE assessment, we employ the evaluation workflow introduced above and report the overall binary classification accuracy across all 179 evaluated criteria for both the \textbf{perturbed} and \textbf{unperturbed} corpora (Table~\ref{tab:results}). Here, ``accuracy'' measures agreement with our LLM-generated and human-reviewed (by the authors of this paper) silver labels, rather than performance against a clinician-adjudicated gold standard.
We test multiple configurations, combining different LLMs, prompting strategies, and two auxiliary \textbf{helper tools} designed to alleviate potential limitations of smaller-scale LLMs. A minimal verification is further conducted on two real tutorial session transcripts to assess consistency with the trends observed on generated data.

\subsection{LLM Selection}\label{subsec:llm_choice}

Due to data privacy concerns, as well as future application as a student training tool, another main focus is on locally deployable open-source models that can be hosted on in-house, reasonably affordable hardware. We thus restrict our selection to quantized models with at most $\sim$32B parameters for computational feasibility: \texttt{qwen3-32b}, \texttt{deepseek-r1-32b}, \texttt{gpt-oss-20b}, \texttt{gemma-3-27b}, \texttt{ministral-8b}, \texttt{llama3.1-8b}, and \texttt{qwen3-8b}, covering both intermediate (20B–32B) and lighter ($\sim$8B) model sizes. For completeness, we also compare these to three state-of-the-art larger LLMs: \texttt{GPT-4o} (used for the data generation earlier; see \autoref{sec:generation}), \texttt{Claude Sonnet~4}, and \texttt{Llama-4-Scout}, to provide higher-end points of reference. While none of these models were specifically trained on French and/or medical terminology, preliminary experiments indicated that most of them display adequate usage of French, and even understanding of most medical terms in the training scenarios. To further investigate this point, we also experimented with injecting medical definitions in-context (see \autoref{subsec:medical_defs}).

\subsection{Prompting Strategies for Evaluation}\label{subsec:prompting_strategy}
We evaluate three prompting strategies. Unlike \citet{geathers2025benchmarkinggenerativeaiscoring}, who use specific prompts for 28 station-generic criteria, we adopt a single prompt template designed to handle all 179 station-specific criteria. 
In the \textbf{zero-shot} setting, the model receives as input the \textbf{transcript}, the \textbf{criterion}, and a \textbf{task description} defining the evaluation objective and output format.
The \textbf{few-shot} setting extends this setup by adding some labeled examples in the task description for guidance \citep{brown2020gpt3}.
\autoref{fig:eval_pipeline} illustrates the inputs and outputs of each criterion evaluation task.
The third strategy is a \textbf{multi-step} variant of the zero-shot setup.
In this configuration, the process is divided into two stages: a first prompt instructs the model to extract from the full transcript the segments most relevant to the target criterion. Then, a second prompt instructs to evaluate this criterion using only the found transcript segments as context.
This design aims to help smaller models maintain focus, as providing the entire transcript may dilute their attention and reduce decision quality.

\subsection{Criterion Decomposition (CD)}\label{subsec:criteria_decomposition}

Many OSCE evaluation criteria are in practice \textbf{composite}, containing explicit logical connectors such as \emph{A \textsc{or} B \textsc{or} C}, \emph{A \textsc{and} B}, or \emph{N \textsc{of} A,B,C}. These formulations can be challenging for LLMs, which may misinterpret their intended logic, for instance by treating an \textsc{or} as an \textsc{and}, or being overly permissive with chained \textsc{and}s. A typical example is the criterion “Asks about the impact of memory (element~A) \textsc{or} concentration problems (element~B),” where the model occasionally assumed that both sub-elements had to be validated simultaneously.
To mitigate such errors, we propose systematically decomposing composite criteria in evaluation sheets into relatively independent sub-criteria by prompting \texttt{GPT-4o} as a preprocessing step. Each sub-criterion obtained is then evaluated as though it were a separate item, and the individual binary decisions are then aggregated programmatically according to the operators, to produce the final decision for the original composite criterion.

\subsection{Medical Definitions (MD)}\label{subsec:medical_defs}

To help the LLMs interpret clinical terminology, we propose to enrich prompts with \textbf{medical definitions}. A medical NER model (\href{https://huggingface.co/medkit/DrBERT-CASM2}{\texttt{medkit/DrBERT-CASM2}}, derived from \citet{labrak2023drbert}) first identifies relevant medical multi-word expressions, which are then matched to concepts in the Unified Medical Language System (UMLS, \citet{UMLS2024AA}), an external biomedical knowledge base. To retrieve definitions for these concepts, we first restrict UMLS to \textbf{French} and \textbf{English} definition sources, as they can be reliably processed by the selected LLMs.
We then select the first available definition for each matched concept from the knowledge base. For example, if “\emph{trouble obsessionnel compulsif}” is identified, it is mapped to the concept \emph{obsessive–compulsive disorder} (OCD) in UMLS, from which a definition such as “An anxiety disorder characterized by ...” is extracted and injected into the prompt.

\begin{table*}[htbp]
\centering
\resizebox{\textwidth}{!}{
\begin{tabular}{lcccccccccccc|cccc}
\toprule
\multirow{2}{*}{\textbf{perturbed}} 
 & \multicolumn{4}{c}{\texttt{qwen3-32b}} 
 & \multicolumn{4}{c}{\texttt{deepseek-r1-32b}} 
 & \multicolumn{4}{c}{\texttt{gpt-oss-20b}}
 & \multicolumn{4}{c}{\texttt{gpt-4o}} \\
\cmidrule(lr){2-5} \cmidrule(lr){6-9} \cmidrule(lr){10-13} \cmidrule(lr){14-17}
 & direct & CD & MD & CD+MD
 & direct & CD & MD & CD+MD
 & direct & CD & MD & CD+MD
 & direct & CD & MD & CD+MD \\
\midrule
zero-shot 
 & 84.36 & \underline{\textbf{90.50}} & 83.24 & \underline{\textbf{90.50}}
 & 83.24 & 84.92 & 82.12 & \textbf{86.03}
 & \textbf{87.71} & 86.59 & \textbf{87.71} & 86.59
 & 84.92 & 87.71 & 84.36 & \textbf{88.83} \\
few-shot  
 & 86.59 & \underline{\textbf{90.50}} & 86.59 & 89.94
 & 82.68 & \textbf{85.47} & 82.12 & 84.92
 & \textbf{87.71} & 84.62 & 86.03 & 84.92
 & 86.03 & \textbf{86.59} & 84.92 & \textbf{86.59} \\
multi-step 
 & 81.01 & \textbf{84.92} & 82.68 & 84.83
 & \textbf{79.89} & 74.86 & 78.77 & 77.65
 & \textbf{83.24} & 78.77 & 81.56 & 78.21
 & 83.80 & \underline{\textbf{85.47}} & 80.45 & 83.24 \\
\bottomrule
\end{tabular}
}

\vspace{0.1cm}

\resizebox{0.80\textwidth}{!}{
\begin{tabular}{lcccccccccccc}
\toprule
\multirow{2}{*}{\textbf{perturbed}} 
 & \multicolumn{4}{c}{\texttt{llama3.1-8b}} 
 & \multicolumn{4}{c}{\texttt{qwen3-8b}} 
 & \multicolumn{4}{c}{\texttt{ministral-8b}} \\
\cmidrule(lr){2-5} \cmidrule(lr){6-9} \cmidrule(lr){10-13}
 & direct & CD & MD & CD+MD
 & direct & CD & MD & CD+MD
 & direct & CD & MD & CD+MD \\
\midrule
zero-shot & 48.04 & 50.84 & 50.84 & \textbf{58.66}
          & 83.24 & 82.12 & 81.56 & \textbf{84.36}
          & 71.51 & \textbf{76.54} & 67.04 & 73.74 \\
few-shot  & 54.75 & \textbf{55.87} & \textbf{55.87} & 54.75
          & \textbf{85.47} & 81.56 & 82.12 & 82.68
          & 62.01 & \textbf{65.36} & 63.13 & 62.57 \\
multi-step& 53.63 & 52.51 & 52.51 & \textbf{56.98}
          & 69.83 & 71.51 & \textbf{72.63} & 69.27
          & 59.78 & 59.78 & 59.22 & \textbf{66.48} \\
\bottomrule
\end{tabular}
}

\vspace{0.2cm}

\resizebox{\textwidth}{!}{
\begin{tabular}{lcccccccccccc|cccc}
\toprule
\multirow{2}{*}{\textbf{unperturbed}} 
 & \multicolumn{4}{c}{\texttt{qwen3-32b}} 
 & \multicolumn{4}{c}{\texttt{deepseek-r1-32b}} 
 & \multicolumn{4}{c}{\texttt{gpt-oss-20b}}
 & \multicolumn{4}{c}{\texttt{gpt-4o}} \\
\cmidrule(lr){2-5} \cmidrule(lr){6-9} \cmidrule(lr){10-13} \cmidrule(lr){14-17}
 & direct & CD & MD & CD+MD
 & direct & CD & MD & CD+MD
 & direct & CD & MD & CD+MD
 & direct & CD & MD & CD+MD \\
\midrule
zero-shot & \underline{\textbf{90.5}} & 89.94 & \underline{\textbf{90.5}} & 89.39
          & 83.8 & 84.36 & \textbf{86.59} & 82.68
          & 86.59 & 83.71 & \textbf{87.71} & 83.8
          & 89.94 & 87.71 & \underline{\textbf{90.5}} & 85.47 \\
few-shot  & \textbf{89.89} & 88.83 & 88.76 & 88.83
          & \textbf{88.83} & 78.77 & 85.47 & 80.45
          & \textbf{84.92} & 83.62 & 83.24 & \textbf{84.92}
          & \underline{\textbf{92.18}} & 87.15 & \underline{\textbf{92.18}} & 88.2 \\
multi-step& \textbf{86.03} & 79.89 & 84.36 & 79.33
          & \textbf{70.95} & 69.83 & \textbf{70.95} & 68.16
          & 69.27 & 73.03 & 70.95 & \textbf{74.3}
          & \textbf{\underline{87.15}} & 79.33 & 86.59 & 81.94 \\
\bottomrule
\end{tabular}
}

\vspace{0.1cm}

\resizebox{0.80\textwidth}{!}{
\begin{tabular}{lcccccccccccc}
\toprule
\multirow{2}{*}{\textbf{unperturbed}} 
 & \multicolumn{4}{c}{\texttt{llama3.1-8b}} 
 & \multicolumn{4}{c}{\texttt{qwen3-8b}} 
 & \multicolumn{4}{c}{\texttt{ministral-8b}} \\
\cmidrule(lr){2-5} \cmidrule(lr){6-9} \cmidrule(lr){10-13}
 & direct & CD & MD & CD+MD
 & direct & CD & MD & CD+MD
 & direct & CD & MD & CD+MD \\
\midrule
zero-shot & 50.84 & \textbf{53.63} & 41.34 & 51.40
          & \textbf{87.64} & 85.39 & 83.24 & 83.24
          & 78.77 & \textbf{84.36} & 78.77 & 78.77 \\
few-shot  & \textbf{54.91} & 54.19 & 53.07 & 51.40
          & \textbf{87.71} & 85.47 & 83.24 & 83.80
          & \textbf{64.80} & 56.42 & 62.01 & 56.98 \\
multi-step& 43.02 & \textbf{49.16} & 39.11 & 44.13
          & \textbf{65.36} & \textbf{65.36} & 60.89 & 61.45
          & 58.66 & 62.01 & \textbf{65.92} & 58.10 \\
\bottomrule
\end{tabular}
}

\caption{\footnotesize Results on both the \textbf{perturbed} and \textbf{unperturbed} corpora. (direct = no auxiliary tools; CD = criterion decomposition; MD = medical definitions). 
For each model, the best accuracy score for each prompting strategy is \textbf{bolded}. The overall best scores across all models are \underline{underlined}. 
gree}
\label{tab:results}
\end{table*}

\section{Results and Discussion}

To assess the feasibility of LLM-based automated OSCE evaluation, we report the overall binary classification accuracy across all 179 evaluated criteria, for both the \textbf{perturbed} and \textbf{unperturbed} corpora (Table~\ref{tab:results}). Some results are omitted for brevity.

\textbf{Overall model performance:} 
Large industry-leading models such as \texttt{GPT-4o} and \texttt{Claude 4 Sonnet} achieve consistently robust performance across all datasets, with accuracies of $\sim$90\% in most configurations. \texttt{Llama-4-Scout} performed slightly less well, possibly on account of its smaller size and Mixture-of-Experts nature.
While previous studies in English OSCE contexts have reported similar observations regarding the promising performance of large models on automated evaluation tasks, those benchmarks are not directly comparable due to linguistic and structural differences. Nonetheless, these results provide encouraging evidence that LLM-based evaluation of French OSCE transcripts is likewise feasible. Though it was used to generate the initial silver labels, some of them were corrected after manual review (see \autoref{sec:labeling}), hence why \texttt{GPT-4o} does not obtain close to perfect scores here.

\autoref{tab:results} shows that models such as \texttt{qwen3-32b}, \texttt{deepseek-r1-32b}, and \texttt{gpt-oss-20b}, together with the lighter \texttt{qwen3-8b}, achieve accuracies comparable to \texttt{GPT-4o}. \texttt{qwen3-32b} is particularly stable, displaying the best accuracy on the \textbf{perturbed} corpus for the \textit{zero-shot} and \textit{few-shot} strategies when using the criterion-decomposition tool, and a close second on the \textbf{unperturbed} corpus. In contrast, \texttt{ministral-8b} displays inconsistent results depending on the strategy and tools used, while \texttt{llama3.1-8b} remains consistently worse in all configurations. Results for \texttt{gemma-3-27b} were excluded because the model failed to consistently produce the required JSON format. These results may reflect differences in these models' training: \texttt{qwen3}, \texttt{deepseek}, and \texttt{gpt-oss} incorporate enhancement methods such as knowledge distillation from their larger variants, while \texttt{llama3.1} and \texttt{ministral} were intentionally fine-tuned with more conventional methods, focusing more on data scale and direct alignment techniques.

\textbf{Effect of prompting strategies:}  
The \textit{zero-shot} setting, while the simplest, proved the most robust, yielding coherent and stable evaluations across cases. Unlike \citet{geathers2025benchmarkinggenerativeaiscoring}, we observe no noticeable degradation with the \textit{few-shot} setting compared to zero-shot. A notable exception was \texttt{ministral-8b}, where few-shot prompting reduced accuracy, possibly due to a sensitivity to prompt length.
In contrast, the \textit{multi-step} strategy degraded performance compared to zero-shot, consistent with prior findings \citep{geathers2025benchmarkinggenerativeaiscoring}. After manual review, we conclude the primary causes were (i) \textit{context loss}, as the evaluation step only sees retrieved partial spans rather than the full transcript, and (ii) \textit{error propagation}, where the retrieval step missing relevant evidence directly biases the downstream binary decision. The latter was especially prevalent, due to the fact criteria-specific information was not necessarily segregated to a few specific spans, but instead often distributed over multiple dialogue turns.

\textbf{Impact of composite-criteria decomposition:}  
The decomposition tool yielded mixed benefits overall. Improvements appeared mainly in the \textbf{perturbed} corpus, where nearly half of the evaluation criteria were transformed into perturbed versions during dialogue generation. Composite criteria (“inquires A \textsc{or} B”) were particularly affected, leading to dialogues in which the virtual doctor executed only one sub-element (“inquires A”) while evaluation remained based on the original combined criterion. Some models (e.g., \texttt{qwen3-32b} and \texttt{llama3.1-8b}) tended to misinterpret \textsc{or} operators as \textsc{and}s, thus mispredicting failures. Decomposition mitigated this by isolating sub-criteria before evaluation. In \textbf{unperturbed} dialogues, where criteria were typically fully executed by the simulated doctor, the effect was more negligible. For \texttt{gpt-oss-20b}, decomposition slightly reduced accuracy, reflecting a tendency towards over-strict evaluation.

\textbf{Impact of medical definition injection:}  
The injection of UMLS-based definitions did not yield clear improvements. This is likely because expert-level medical terminology was infrequent, most models already understood common medical terms, and the identification of multi-word medical expressions introduced additional noise. Future gains may come from more selective filtering and evaluations on more jargon-heavy OSCE scenarios.

\textbf{Preliminary verification on real cases:}
Previous experiments on synthetic transcripts confirmed that several models were sufficiently performant and stable for automated OSCE evaluation. To verify whether these findings extend to real data, we conducted an additional small-scale experiment using two authentic transcripts (38 criteria in total) obtained after-the-fact from tutorial OSCE sessions, where a teacher played the role of the doctor (\autoref{tab:feasibility}). The table reports results for a representative subset of models: the reference model \texttt{GPT-4o}, the most performant and stable open-source model \texttt{qwen3-32b}, the lighter yet competitive \texttt{qwen3-8b}, and the weaker baseline \texttt{llama3.1-8b}. Results followed the same overall trends observed on synthetic data, with accuracy levels remaining consistent across settings. 
Although limited, this small-scale experiment suggests that the relative trends observed on synthetic transcripts may carry over to these tutorial dialogues.

\begin{table}[h!]
\centering
\resizebox{\columnwidth}{!}{
\begin{tabular}{lcccc}
\toprule
\textbf{Real case} & qwen3-32b & gpt-4o & qwen3-8b & llama3.1-8b \\
\midrule
Zero-shot & 86.97 & 83.47 & 75.77 & 37.25 \\
\bottomrule
\end{tabular}
}
\caption{Average zero-shot accuracy on two real-case OSCE transcripts, without helper tools.}
\label{tab:feasibility}
\end{table}

\textbf{Discussion:}
Our experiments suggest that general-purpose LLMs show strong potential for automated evaluation in French OSCE settings. While there were initial concerns about their ability to handle French medical terminology, the results indicate that this limitation may be less severe than expected: for most common clinical terms encountered in OSCEs, current models appear encouraging, especially compared to the currently purely indicative grading used in OSCE training sessions, due to the lack of expert examiners.
Although the \textit{preliminary verification on real transcripts} exhibited trends similar to those observed on synthetic data, a more systematic validation of the synthetic dataset remains necessary once larger sets of authentic transcripts become available. In particular, stricter validation should compare generated and real dialogues directly—for instance, by aligning criterion labels and applying similarity metrics such as BLEURT \citep{sellam-etal-2020-bleurt}, or by using \textit{LLM-as-a-judge} approaches.

\section{Conclusion}

In this work, we developed a controlled pipeline for generating synthetic French OSCE training transcripts alongside an automated evaluation framework for clinical-skills criteria based on locally hostable LLMs. By structuring dialogue generation around evaluation criteria and incorporating perturbations to simulate less idealized student performances, we produced two small datasets for benchmarking LLM-based automated evaluation.

Our experiments indicate that mid-size open-source LLMs (20–32B parameters) can deliver performance comparable to much larger language models, while lighter models ($\sim$ 8B) remain competitive and may even support closer-to-real-time usage. These findings are encouraging, and highlight the feasibility of locally deployable solutions, that would ensure greater control and respect for privacy in this kind of educational setting. 
In our binary evaluation task, we also found that although not specifically trained on the medical domain, recent general-purpose LLMs generally handle both French and its medical terminology well, making external definition injection unnecessary in most cases. However, their handling of composite criteria and logical operators remains less reliable, meaning that supportive external tools are still necessary for some subtasks.

While encouraging, important challenges remain. The synthetic transcripts generated with this framework are likely to reflect more idealized student performances, highlighting the need for larger-scale validation against authentic OSCE data with medical expert annotations. In addition, our evaluation focused primarily on binary labels: future studies should also assess the quality of generated justifications and the relevance of transcript excerpts provided as evidence. 
In future work, we plan to expand the diversity of generated scenarios to better reflect the breadth of OSCEs.
Medical expert educators will also be involved to calibrate the evaluation strictness modes and validate the automatic labeling of generated transcripts. 
\section*{Acknowledgements}
We would like to thank Yongqiang Yu for proposing the initial idea of adjustable levels of strictness in the evaluation framework and for insightful discussions during the development of this work.

\section*{Data and Code Availability Statements}

Experiments with local LLMs were performed with a mixture of locally hosted models and, for convenience, OpenRouter-hosted models to parallelize runs, using \texttt{Q4\_K\_M} quantizations as a balance between resource efficiency and quality. All prompts, generated transcripts, and implementation details used in this work will be released as supplementary material upon acceptance.

\section*{Limitations}
The generated OSCE dialogues, while diverse, remain partly idealized because explicit evaluation criteria guide their structure. As a result, they tend to under-represent spontaneous or ``off-script'' behaviors typical of real student performances. 
In particular, the generated dialogues are well-structured and logically coherent but lack hesitations, repetitions, self-corrections, interruptions, and occasional off-topic responses that are common in authentic OSCE conversations.
Perturbations of leaf criteria only partially mitigate this effect, and the resulting errors remain comparatively ``structured'' rather than chaotic, compounding, or unstable as in real student behavior.
 Larger validation sets, including authentic transcripts with expert annotations, are therefore needed to assess ecological validity; our real-data check is limited to two tutorial-session transcripts (38 criteria) and should not be viewed as a full validation.

Additionally, our synthetic dialogue transcripts do not reproduce the transcription artifacts and conversational noise typically present in real recordings (e.g., disfluencies, interruptions, or incomplete turns). As such, the present study remains theoretical with respect to the robustness of evaluation methods to real-world transcription variability.
The silver-labeling procedure used \texttt{GPT-4o} to produce reference decisions without expert intervention.
Although the labels were manually reviewed for consistency, they did not undergo a formal adjudication process. Accordingly, the labels should be considered a reviewed silver standard rather than definitive ground truth, and the reported accuracies reflect agreement with this reference rather than clinician-adjudicated validity.

Finally, since the synthetic transcripts were generated with \texttt{GPT-4o}, they may reflect its linguistic style, which could bias results and partially inflate evaluation performance for models that align better with this style.
Future work can mitigate this risk by diversifying generation (e.g., multiple generators / style variation) and by evaluating on larger sets of authentic transcripts when available.

A single evaluation run was performed for each configuration. A temperature of zero was not used because several models still produced non-deterministic and/or incoherent outputs under that setting; instead, following preliminary experiments, a temperature of 0.2 was adopted, balancing determinism and output quality. While averaging over multiple runs could further improve robustness, this was not pursued due to time and resource constraints. Nevertheless, the current results provide a sufficient basis for a first exploration of the feasibility of LLM-based evaluation in French OSCEs.

Finally, the evaluation pipeline focused exclusively on binary decisions. Other aspects, such as the linguistic adequacy of student utterances, non-verbal behavior, or the quality and pedagogical value of model-generated justifications and extracted evidence segments, remain outside the present scope, but may be explored in future work.

\section*{Ethics Statement}

This study did not involve any collection or processing of personal, clinical, or patient-identifiable data. All dialogues used for experimentation were synthetically generated by large language models (LLMs) based solely on OSCE training scenarios invented by medical educators. No recordings or transcripts from real examinations or students were used at any stage, other than purely as inspiration for defining evaluation standards and understanding the pedagogical context of such practical exams.

Because all data are artificial, the system presented in this paper is not intended for immediate deployment in real training or examination settings. While our approach provides a controlled framework for studying LLM-based assessment, the synthetic nature of the dialogues implies that unknown model biases may affect both the realism of generated interactions and the reliability of automated judgments. Furthermore, the justifications and evidence segments produced by the models have not yet been systematically evaluated for their appropriateness or pedagogical value; such verification is planned for future research. Accordingly, these outputs should not be used as educational feedback without prior expert validation and ethical oversight. Consequently, any educational use of the proposed system would require prior expert validation, bias auditing, and alignment with ethical guidelines governing medical training and evaluation.

\section*{Bibliographical References}\label{sec:reference}

\bibliographystyle{lrec2026-natbib}
\bibliography{master}

\end{document}